\documentclass[letterpaper]{article} 
\usepackage{aaai2026}  
\usepackage{times}  
\usepackage{helvet}  
\usepackage{courier}  
\usepackage[hyphens]{url}  
\usepackage{graphicx} 
\usepackage{natbib}  
\usepackage{caption} 
\usepackage{multicol}
\usepackage{multirow}
\usepackage{booktabs}
\usepackage{enumitem}
\usepackage{amsthm}  
\newtheorem{definition}{Definition}
\newtheorem{lemma}{Lemma}
\usepackage{algorithm}
\usepackage{algorithmic}
\usepackage{subcaption}
\usepackage{amsfonts}
\usepackage{tabularx}
\usepackage{amsmath}
\usepackage{newfloat}
\usepackage{listings}
\DeclareCaptionStyle{ruled}{labelfont=normalfont,labelsep=colon,strut=off} 

\usepackage{xcolor}         
\usepackage{xspace}
\usepackage[capitalize,nameinlink]{cleveref}   

\floatstyle{ruled}
\newfloat{listing}{tb}{lst}{}
\floatname{listing}{Listing}
\title{The Easy Path to Robustness: Coreset Selection using Sample Hardness}
\author{
    Pranav Ramesh \textsuperscript{\rm 1}, 
    Arjun Roy \textsuperscript{\rm 2}, 
    Deepak Ravikumar \textsuperscript{\rm 2}, 
    Kaushik Roy \textsuperscript{\rm 2}, 
    Gopalakrishnan Srinivasan \textsuperscript{\rm 1}
}
\affiliations{
    \textsuperscript{\rm 1}Indian Institute of Technology, Madras\\
    \textsuperscript{\rm 2}Purdue University\\

}

\newcommand{\sys}{EasyCore\xspace}

\usepackage{bibentry}

\begin{document}

\maketitle

\begin{abstract}
Designing adversarially robust models from a data-centric perspective requires understanding which input samples are most crucial for learning resilient features. While coreset selection provides a mechanism for efficient training on data subsets, current algorithms are designed for clean accuracy and fall short in preserving robustness. To address this, we propose a framework linking a sample's adversarial vulnerability to its \textit{hardness}, which we quantify using the average input gradient norm (AIGN) over training. We demonstrate that \textit{easy} samples (with low AIGN) are less vulnerable and occupy regions further from the decision boundary. Leveraging this insight, we present \sys, a coreset selection algorithm that retains only the samples with low AIGN for training. We empirically show that models trained on \sys-selected data achieve significantly higher adversarial accuracy than those trained with competing coreset methods under both standard and adversarial training. As AIGN is a model-agnostic dataset property, \sys is an efficient and widely applicable data-centric method for improving adversarial robustness. We show that \sys achieves up to 7\% and 5\% improvement in adversarial accuracy under standard training and TRADES adversarial training, respectively, compared to existing coreset methods. 
\end{abstract}    
\section{Introduction}
\label{sec:intro}

Deep Neural Networks have been extremely successful in performing complex tasks across a wide range of domains, such as computer vision and natural language processing~\cite{lecun2015deep}. Despite their extensive adoption, neural networks have been shown to be vulnerable to adversarial attacks, wherein the attacker adds a small adversarial noise to the input where this range of noise is defined as bounded pixel-space $\epsilon$ balls. This noise is indiscernible to humans, but tricks the neural network to misclassify the image. 

In order to mitigate the effect of such adversarial examples, the technique of adversarial training was suggested. This helps in mitigating against adversarial attacks, but also reduces the clean accuracy of the classifier (accuracy on unperturbed test set). Adversarial training is expensive, increasing the training time by orders of magnitude. To address this concern, adversarial training methodologies were developed which reduced training time~\cite{shafahi2019adversarialtrainingfree, wong2020fastbetterfreerevisiting}. These methods established that adversarially robust models could be trained using a weaker and cheaper adversary. 

However, a more fundamental drawback of standard training is its sensitivity to well-generalizing features in the data. It was found that standard machine learning datasets show certain non-robust features, which are learned by models ~\cite{ilyas2019adversarialexamplesbugsfeatures, schmidt2018adversarially}. This has motivated a shift towards a data-centric perspective by balancing between generalization and memorization of networks. This perspective focuses on the effect of prototypical and atypical samples on memorization \cite{feldman2021doeslearningrequirememorization, feldman2020neural, ravikumar2024unveilingprivacymemorizationinput}. In this work, we use the data-centric objective to identify a smaller, computationally efficient dataset subset, or coreset, that is robust to adversarial attacks. This is in contrast to several existing coreset selection methods \cite{mirzasoleiman2020coresetsdataefficienttrainingmachine, xia2024refinedcoresetselectionminimal, sener2018activelearningconvolutionalneural, xia2023moderate, Zheng2022CoveragecentricCS}, which are typically designed for clean accuracy and often fail to retain the specific samples that are crucial for learning a robust classifier.

We show that the key to an adversarially robust coreset lies in understanding the extent to which each sample's features are represented during the course of training. We quantify this notion using \textbf{Average Input Gradient Norm (AIGN) over training} (see \Cref{def:AIGN}). AIGN is defined as the average of the magnitude of input gradient norm of a sample averaged over training. AIGN serves as a powerful and efficient proxy for how well-learned a particular sample is \cite{ravikumar2025towards}. This notion of AIGN is intrinsically linked to adversarial vulnerability, i.e. the adversary's ability to push the $\epsilon-$ball surrounding the sample across the decision boundary to an adjacent class \cite{madry2019deeplearningmodelsresistant}. This is because adversaries also use the direction of the input gradient norm to find adversarial examples. We describe the links between AIGN, decision boundary curvature and adversarial robustness in more detail in later sections. 

\begin{definition}
The AIGN for a sample $\vec{x}$, AIGN($\vec{x}$) = $\frac{1}{N} \sum_{i=1}^{N} ||\nabla_{\vec{x}}^i l(\vec{x}, y)||_2$, where $N$ is the number of epochs, $y$ is the label, $l(\cdot)$ is the loss function, and $||\cdot ||_2$ denotes the norm. 
\label{def:AIGN}
\end{definition} 

\begin{figure}[tb!]
\centering
\begin{subfigure}{0.23\textwidth}
    \includegraphics[width=\textwidth]{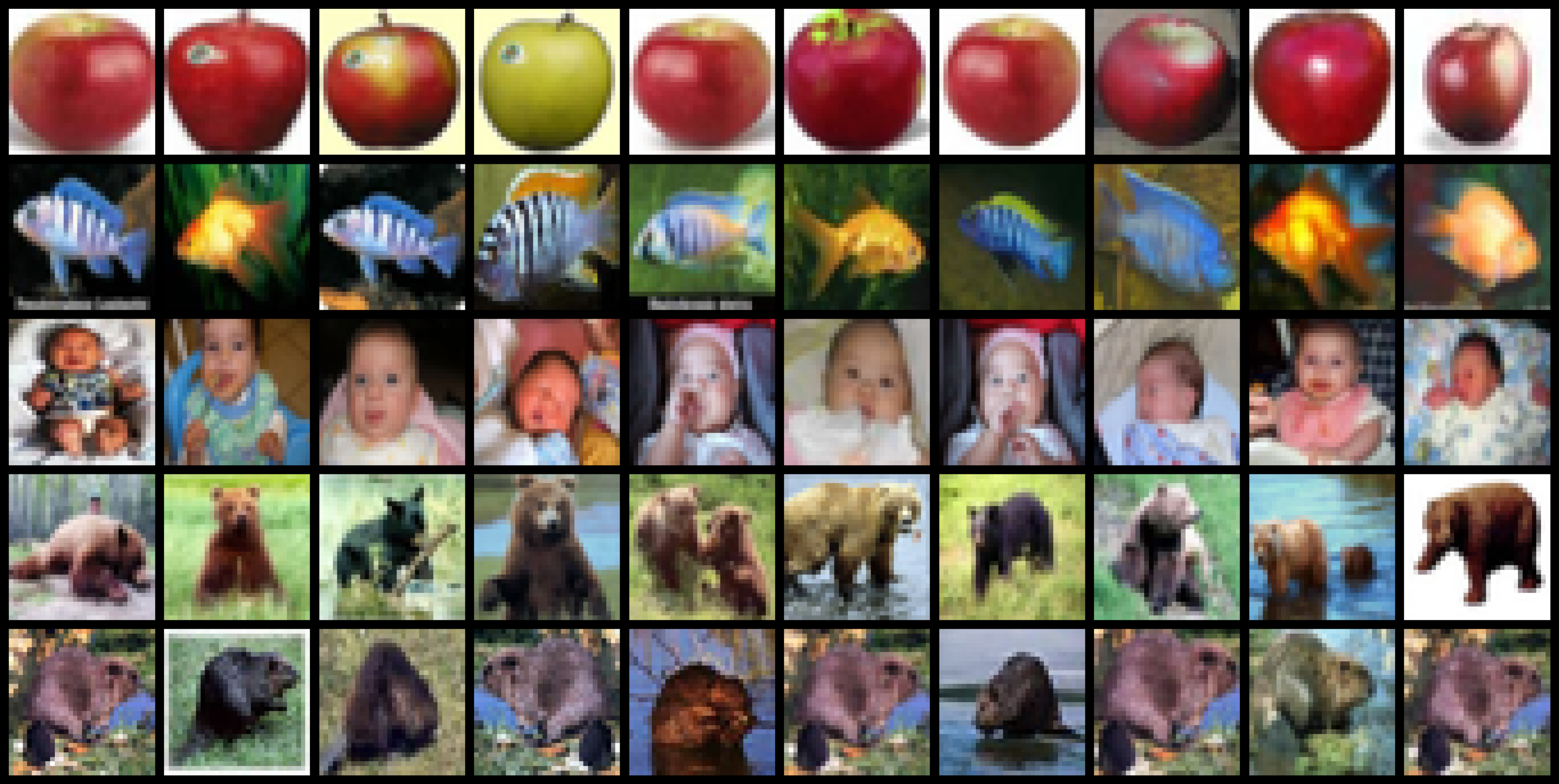}
    \caption{}
\end{subfigure}
\hfill
\begin{subfigure}{0.23\textwidth}
    \includegraphics[width=\textwidth]{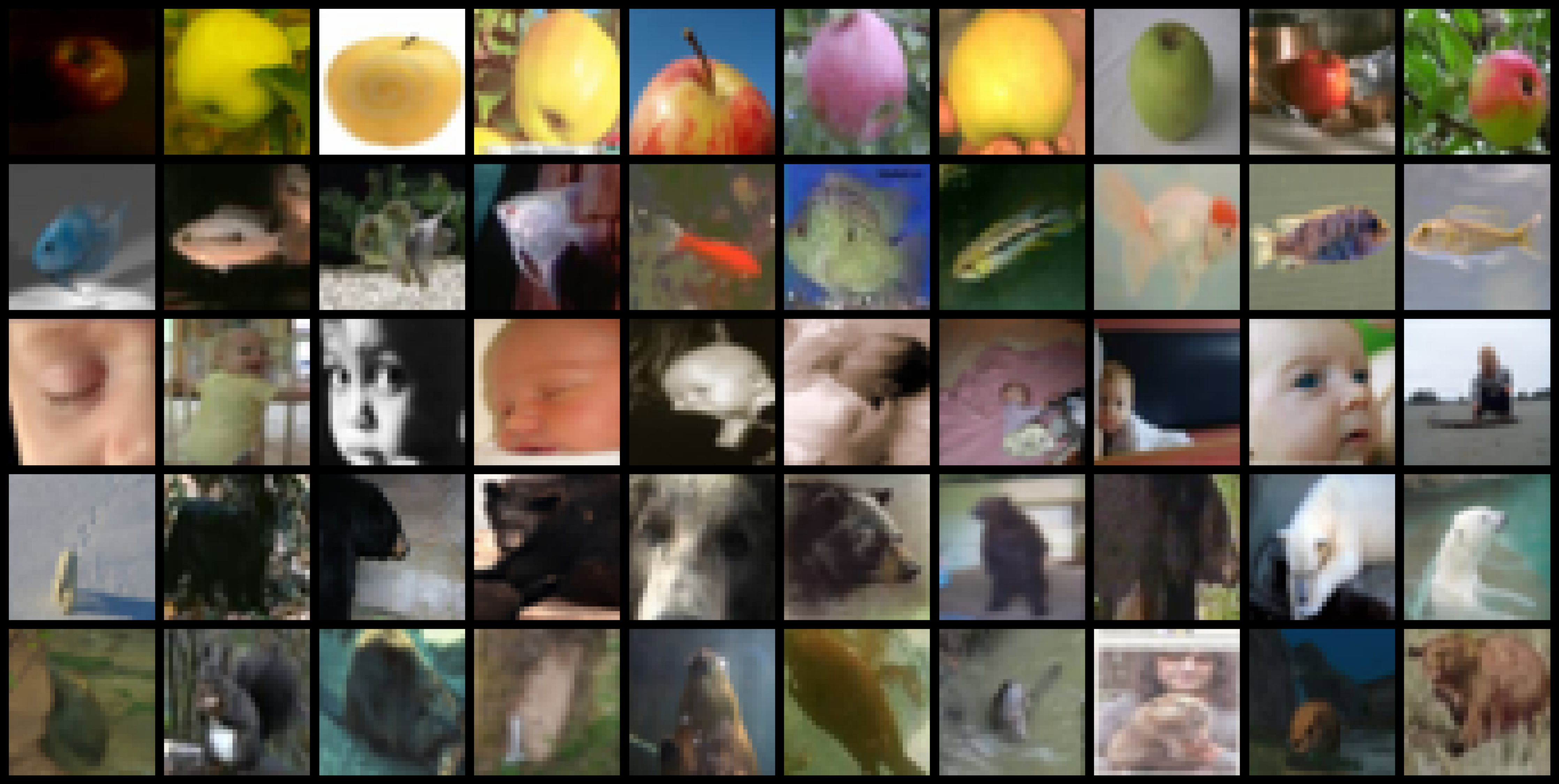}
    \caption{}
\end{subfigure}
\caption{(a) the 10 \textit{easiest} examples (least AIGN) and (b) the 10 \textit{hardest} examples (highest AIGN) of the same class, for five CIFAR-100 classes}
\label{fig:easy_hard_cifar100}
\end{figure}

Sample hardness is an important indicator to understand how easily a model can \textit{learn} the sample. Samples can be hard for multiple reasons, such as mislabeling and atypical-ness. In the rest of this paper, we use sample hardness to qualitatively describe the variation of AIGN within samples. In particular, we define the \textit{hardness order} of a dataset in \Cref{def:AIGN_easy}. This definition is motivated further in the Methodology section of the paper. 
We visualize the easiest and hardest examples of a few classes in the CIFAR-100 dataset in \Cref{fig:easy_hard_cifar100}. 

\begin{definition}
The \textbf{hardness order} of a dataset is same as the samples ordered in \textbf{ascending order} of their AIGN. In other words, \textbf{easy} samples are those which have a relatively \textbf{lower} AIGN, whereas \textbf{hard} examples have a relatively \textbf{high} AIGN. 
\label{def:AIGN_easy}
\end{definition} 

We observe that easy samples are learned quickly and exhibit consistently low input gradient norms, while hard samples maintain high gradients for longer and often reside near highly curved regions of the decision boundary, making them more susceptible to attacks.

Based on this insight, we introduce \textbf{\sys}, a simple yet effective methodology for building adversarially robust coresets. \sys prioritizes the selection of easy samples over hard samples. By training on these samples, our method encourages a smoother, more regularized decision boundary, leading to improved adversarial robustness even at high data pruning rates. As an added benefit, because the ranking of samples by input gradient is a stable dataset property \cite{Garg_2023_CVPR}, AIGN needs to only be computed once for a given dataset even by using other networks \cite{Coleman2020Selection}. This allows the \sys method to be both efficient and model-agnostic, making it a practical solution for scalable robust training. Initially, we demonstrate robustness benefits under standard training, thereby highlighting the efficacy of our data-centric approach. Subsequently, we also benchmark \sys under adversarial training, where we continue to show robustness improvements over competing methods. In this paper, our key contributions are as follows: 

\begin{itemize}
    \item We propose \textbf{\sys}, a coreset selection technique designed to improve adversarial robustness by focusing on training data distribution and sample hardness. Since \sys is a \textit{data-centric} approach, it is model-agnostic, and requires analyzing the average input gradient norm over training, of a single sample model.
    \item We show that \sys significantly improves adversarial accuracy on a wide range of datasets and coreset fractions in comparison with existing approaches. We show that \sys achieves up to 7\% and 5\% improvement in adversarial accuracy under standard training and TRADES~\cite{zhang2019trades} adversarial training, respectively, compared to previous coreset methods. 
    \item We investigate the link between AIGN, prototypical-ness of samples, and decision boundaries using various visualization techniques to enhance the explainability of \sys. We also study the effect of adversarial training on each of the above metrics. 
\end{itemize}

\begin{figure*}
\includegraphics[width=\textwidth]{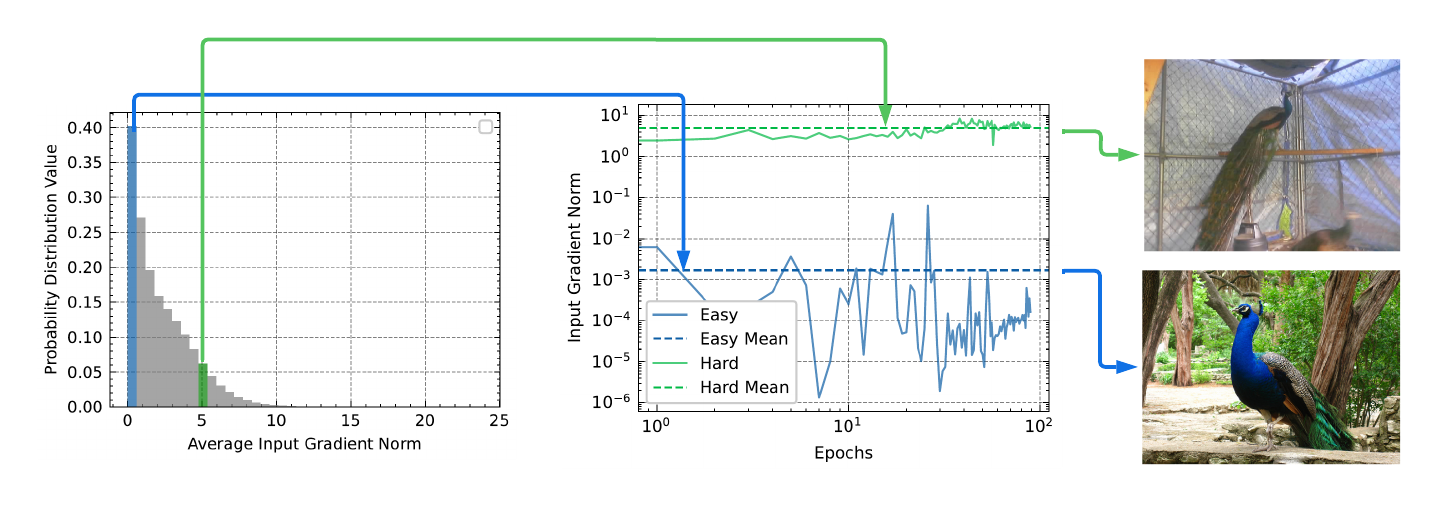}
\caption{Visualization of easy and hard samples from the ImageNet \textit{peacock} class, identified by their average input gradient norm (AIGN). Left: The long-tailed distribution of AIGN scores. Middle: The input gradient norm training trajectory for easy prototypical samples with low AIGN (blue) and hard atypical samples with high AIGN (green). Right: The easy image (bottom) is typical of the class, whereas the hard image (top) is more atypical of the class, and thus is harder to learn.}
\label{fig:intuition}
\end{figure*}

\section{Background and Related Work}
\label{sec:background}

\textbf{Coreset Selection}: The main objective of coreset selection is to identify a small subset of a large dataset that can approximate the performance of the full set, thereby accelerating model training. This problem has been approached from various angles, with most methods designed to optimize for standard model accuracy on clean data. These approaches often select points by matching full-dataset gradient information \cite{mirzasoleiman2020coresetsdataefficienttrainingmachine}, assessing dataset geometry \cite{sener2018activelearningconvolutionalneural, xia2023moderate, Zheng2022CoveragecentricCS}, or identifying uncertain samples \cite{xia2024refinedcoresetselectionminimal}. In contrast, our work, \sys, introduces a new selection criterion, sample \textit{hardness} via AIGN, specifically to construct coresets that enhance adversarial robustness, a goal for which standard coreset methods are not optimized for.

Adversarial Coreset Selection (ACS) \cite{hadi} is a method that explores coreset selection for adversarial robustness. They formulate a robust coreset selection as a bi-level optimization problem, using greedy solvers during training. This creates a \textit{dynamic} coreset during training, whereas \sys uses a simple \textit{static} coreset selection criterion computed once before training (see the Results section for further detailed comparison).



\textbf{Scoring and Ordering of Samples}: Many methods leverage scores to rationalize the training process. \textbf{Importance sampling}, for instance, uses scores to define a non-uniform sampling distribution, typically re-weighting gradients to maintain an unbiased estimate \cite{imp1, imp2, imp3}. \textbf{Curriculum learning} uses a similar notion of sample importance to create a meaningful ordering, often proposing an ``easy-to-hard" progression to improve model convergence and generalization \cite{bengio2009curriculum, jiang2014self}. This \textit{easy-vs-hard} dichotomy is also central to understanding model \textbf{memorization}, where \textit{hard} samples are often atypical examples that a model is forced to memorize, while \textit{easy} samples correspond to generalizable, prototypical features \cite{feldman2021doeslearningrequirememorization, feldman2020neural, ravikumar2024unveilingprivacymemorizationinput}. \sys synthesizes these perspectives into a unified, \textit{data-centric} framework for robust coreset selection by using a single score, the Average Input Gradient Norm (AIGN), to quantify its likelihood of being memorized versus generalized~\cite{ravikumar2025towardsthesis}.

\textbf{Achieving Robustness via Algorithmic and Geometric Perspectives}: The most direct method for improving adversarial robustness is \textbf{adversarial training}, an algorithmic approach that augments the training set with worst-case examples generated via attacks like projected gradient descent (PGD) \cite{madry2019deeplearningmodelsresistant} and fast gradient sign methods (FGSM) \cite{fgsm}. While effective, this process often results in a trade-off between clean accuracy, adversarial robustness and computational efficiency \cite{zhang2019trades, zhao2023fast}. From a geometric perspective, the goal of these methods is to create a more favorable \textbf{decision boundary}. It is widely hypothesized that smoother, less curved decision boundaries are inherently more robust, as small input perturbations are less likely to cross into an incorrect class region \cite{Lei2025understanding, yang2024refined}. Prior work suggests that this desirable geometry can be achieved by focusing on well-represented, prototypical samples during training \cite{carlini2019distributiondensitytailsoutliers}. By selecting a coreset of \textit{easy} samples (low AIGN), \sys provides a practical and efficient mechanism for encouraging a smoother decision boundary, improving adversarial robustness both standard and adversarial training.

\section{Methodology}
\label{sec:methodology}

This section outlines the intuition and reasoning behind the development of our adversarial coreset selection methodology. To achieve this, we first describe an example to build intuition regarding the understanding of input gradient norm. Next, we establish the link between input sample gradient and adversarial vulnerability \cite{ravikumar2024unveilingprivacymemorizationinput, Moosavi-Dezfooli_2019_CVPR}. Finally, we describe our coreset selection methodology, leveraging the Average Input Gradient Norm (AIGN) as a simple score for the development of our adversarial coreset selection method \sys.

\subsection{Motivating Example}
Optimization theory often provides proofs on convergence with respect to the norm of the weight gradients \cite{ghadimi2013stochastic}. That is, as a network is trained, the gradient with respect to its weights will converge. Interestingly, research \cite{imp2, ravikumar2025towards} has shown a link between the input gradient and the weight gradient, showing input gradients also converge during training.

We make an additional observation: the average input gradient norm over the course of training can effectively capture sample hardness. Figure \ref{fig:intuition} illustrates this by visualizing the input gradients of two ImageNet images, one categorized as \textit{easy} and the other as \textit{hard}. The gradients for \textit{easy} samples initially decrease and stay low, suggesting they are quickly learned. In contrast, the gradients for \textit{hard} samples remain high for much longer through the course of training. Given the inherent noise in gradient calculations, averaging the input gradients provides a robust and effective method for measuring how difficult a particular sample is for the model to learn.

Sample hardness is strongly related to decision boundary geometry and adversarial vulnerability \cite{Moosavi-Dezfooli_2019_CVPR, carlini2019distributiondensitytailsoutliers}. Figure \ref{fig:u_curve_ablation} provides evidence for this link between sample hardness, captured by the input sample gradient, and adversarial vulnerability. Therefore, capturing sample hardness via the input sample gradient serves as a good proxy to identify sample vulnerability.

\subsection{Coreset Selection using \sys}
The development of \sys draws inspiration from the two insights above: (a) input gradient captures sample hardness, and (b) input gradient is linked to adversarial vulnerability and the corresponding decision boundary geometry.

The intuition for building an adversarial coreset is simple: if we capture and use the \textit{easy} subset that is, samples that are adversarially less vulnerable as shown above, this results in a model with a flatter decision boundary, which possesses better adversarial robustness.
A smoother boundary results in non-boundary \textit{hard} samples being further from the boundary, thereby increasing their adversarial distances. This approach, however, necessitates a careful balance, as prioritizing robust-critical samples may come at the cost of clean accuracy. 



\begin{figure}[h]
\centering
    \includegraphics[width=0.35\textwidth]{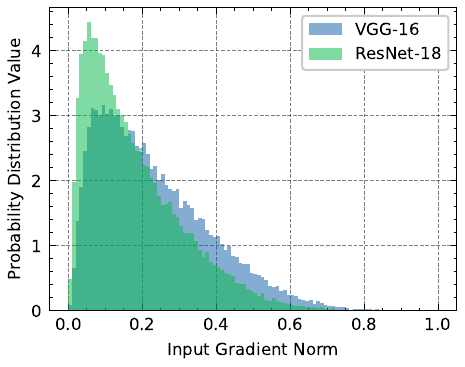}
    \caption{Comparison of normalized average input gradient norm on CIFAR-10 between a VGG-16 and ResNet-18 model}
    \label{fig:grad_in_norm_vggvsresnet}
\end{figure}

The \sys coreset selection algorithm operationalizes this intuition by computing the input gradient norms of samples through training.
It then selects the samples with the lowest average input gradient norm as the coreset selection methodology.
The overall procedure is described in \Cref{alg:EasyCore}. 
\begin{algorithm}
    \caption{\sys}
    \label{alg:EasyCore}
    \begin{algorithmic}[1]
        \STATE \textbf{Input:} Dataset $\mathcal{D}$, coreset fraction $f$
        \STATE Train a fixed model $\mathcal{M}$ for $e$ epochs, obtaining epoch-wise checkpoints $C_i$ for $i \in [0,e)$. 
        \STATE Compute $||\nabla^i_xl(x,y)||_2$ for each checkpoint, and use that to compute $\mathbb{E}_{epochs}[||\nabla_xl(x,y)||_2]$. 
        \STATE Sort the input images by their values of $\mathbb{E}_{epochs}[||\nabla_xl(x,y)||_2]$ to get a list of input indices $arr$
        \STATE \textbf{return} The first $f \cdot |X_t|$ values in $arr$, where $|X_t|$ is the size of the training set
    \end{algorithmic}
\end{algorithm}

\section{Discussion}
We now further reinforce the link between Average Input Gradient Norm (AIGN), prototypical-ness of samples and decision boundary curvature. Subsequently, we also analyze the effect of adversarial training on the model's ability to learn features. In effect, we make the following claims: 

\textbf{Claim 1: Lower AIGN indicates prototypical-ness} Samples with lower AIGN are \textit{prototypical} to the class, i.e. they constitute the set of samples with features best defining their class.

\textbf{Claim 2: Training on high-AIGN examples increases decision boundary curvature}: Samples with lower AIGN are further from the decision boundary between classes, and training with samples having higher AIGN results in a more curved decision boundary. 

\textbf{Claim 3: Training on harder samples reduces robustness on easy samples}: Training on samples with higher AIGN (\textit{harder} samples) results in degraded adversarial robustness on the core easy samples. In other words, \sys performs better than a uniformly chosen coreset on easy samples. Also, training on easy train samples makes the model adversarially robust on easy test samples. 

\textbf{Claim 4: Adversarial training uses more dimensions to learn features}: Adversarial training makes the penultimate layer of the classifier less amenable to dimensionality reduction, and thus more components are required to capture 95\% of the variance. This in turn means that the model \textit{consumes} more of its available parameter space in order to be adversarially robust. As a result, the peak normalized AIGN shifts to the right. We substantiate each of these claims below. 
\begin{figure}[h!]
\centering
\begin{subfigure}{0.3\textwidth}
    \includegraphics[width=\textwidth]{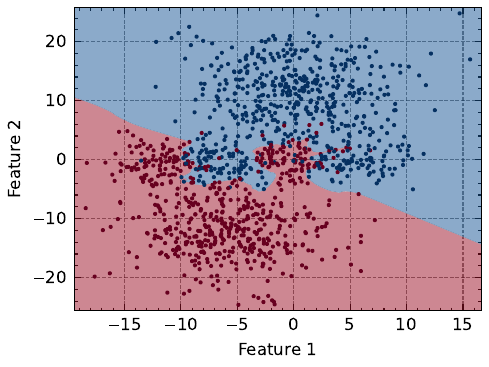}
    \caption{}
    \label{fig:DRI_vs_decision_boundary_1}
\end{subfigure}
\hfill
\begin{subfigure}{0.3\textwidth}
    \includegraphics[width=\textwidth]{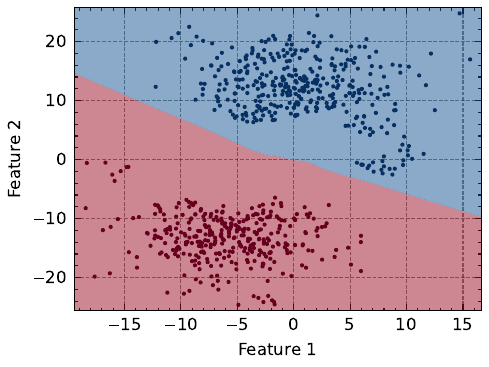}
    \caption{}
    \label{fig:DRI_vs_decision_boundary_2}
\end{subfigure}
\caption{A two-class dataset comprising of 1200 points in 2D space. (a) Decision boundary and training points of the entire dataset (b) Decision boundary of a model trained on the \textit{easiest} 720 points (ordered by AIGN), and the corresponding data points. It can be seen that the decision boundary is more curved and complex in (a) compared to (b). }
\label{fig:DRI_vs_decision_boundary}
\end{figure}

\begin{figure*}[h!]
\centering
\begin{subfigure}{0.2\textwidth}
    \includegraphics[width=\textwidth]{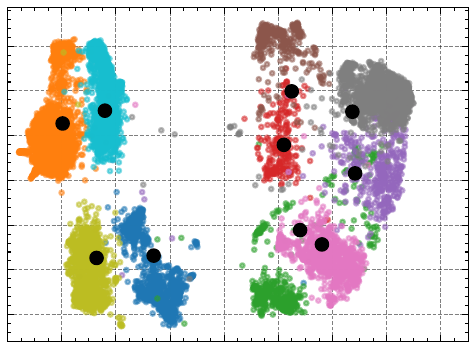}
    \caption{}
    \label{fig:UMAP_plots_1}
\end{subfigure}
\hfill
\begin{subfigure}{0.2\textwidth}
    \includegraphics[width=\textwidth]{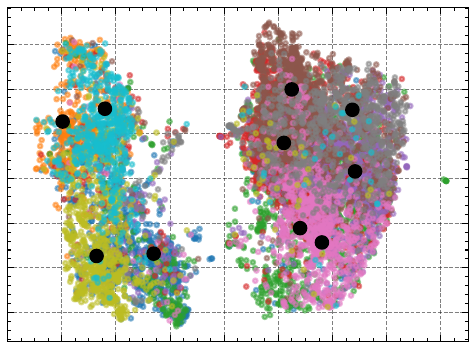}
    \caption{}
    \label{fig:UMAP_plots_2}
\end{subfigure}
\hfill
\begin{subfigure}{0.2\textwidth}
    \includegraphics[width=\textwidth]{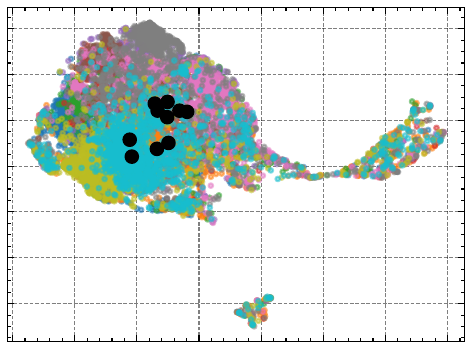}
    \caption{}
\end{subfigure}
\hfill
\begin{subfigure}{0.2\textwidth}
    \includegraphics[width=\textwidth]{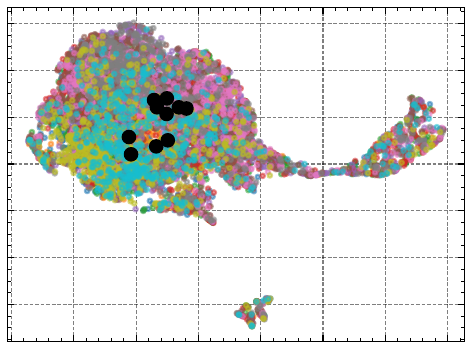}
    \caption{}
\end{subfigure}
\caption{Scatter plot of 10000 images with (a) least AIGN (\textit{easy)} and (b) highest AIGN (\textit{hard}) \textbf{under standard training}, and (c) least AIGN (\textit{easy)} and (d) highest AIGN (\textit{hard}) \textbf{under adversarial training}, for the CIFAR-10 dataset. Points marked in black denote the centroids of each class label. Points are coloured as per their class labels. It can be seen that the easy images form the core cluster around the centroid for each class, and thus can easily be differentiated compared to hard images. }
\label{fig:UMAP_plots}
\end{figure*}

\subsection{Claim 1: Lower AIGN indicates prototypical-ness}
\label{sec:UMAP_viz}
In order to establish that samples with lower input gradient norm are well-learned, we perform dimensionality reduction in the feature vector space at the penultimate layer of the model. This results in a 2D view of data, which can be used for visualization. We use UMAP, which is a well known general-purpose dimensionality reduction technique for machine learning~\cite{mcinnes2020umapuniformmanifoldapproximation}. We show that average input gradient norm captures representation at the penultimate layer well. We do this by observing the points corresponding to the 10000 \textit{easiest} images as per \Cref{def:AIGN_easy} in comparison with the 10000 \textit{hardest} images (see \Cref{fig:UMAP_plots} (a),(b)).  Using this, we can see that images with lower AIGN form the center of each label cluster in lower-dimension space, thus characterizing the primary features of the class. Images with higher AIGN, on the other hand, reside on the fringes of each class in lower-dimension space and intermix with the features of other classes.

\subsection{Claim 2: Training on high-AIGN examples increases decision boundary curvature}

To understand the effect of the \textit{hardness} of samples on the decision boundary curvature, we consider a simple 2D dataset comprising of gaussian points at different centroids, with only two classes (see \Cref{fig:DRI_vs_decision_boundary}). We use a sufficiently parametrized model with 20 residual blocks, each with feature dimension 256, for an overall parameter count of 1317.1K. We compare two scenarios, \Cref{fig:DRI_vs_decision_boundary_1} shows training on the complete dataset and \Cref{fig:DRI_vs_decision_boundary_2} shows training only on easier samples (refer \Cref{def:AIGN_easy}). As the model is sufficiently parametrized to fit the various clusters, we can see that the decision boundary curvature increases around the harder samples, hence the more complex boundaries in \Cref{fig:DRI_vs_decision_boundary_1}. This then leads to the fact that training on the entire dataset reduces the adversarial robustness of the model on the easy samples, due to the increased curvature of the decision boundary. This shows the validity of claim 3 as well on a simple dataset. However, in the next section, we provide results for CIFAR-10~\cite{krizhevsky2009cifar}, a real-life dataset. 

\begin{table*}[h!]
    \centering
    \caption{Accuracy under standard training for different methods and coreset fractions on CIFAR-10 and CIFAR-100. EasyCore performs better than standard coreset methods in terms of adversarial accuracy across coreset fractions. For CIFAR-100, at 60\% coreset size, EasyCore achieves over 7\% improvement in adversarial accuracy at comparable clean accuracy. }
    \label{tab:accuracy_results_combined}
    \resizebox{\textwidth}{!}{
   \begin{tabular}{llcccccccc}
    \toprule
    \textbf{Dataset} & \textbf{Method / Fraction} 
    & \multicolumn{2}{c}{\textbf{0.05}} 
    & \multicolumn{2}{c}{\textbf{0.20}} 
    & \multicolumn{2}{c}{\textbf{0.40}} 
    & \multicolumn{2}{c}{\textbf{0.60}} \\
    \cmidrule(lr){3-4} \cmidrule(lr){5-6} \cmidrule(lr){7-8} \cmidrule(lr){9-10}
    & & \textbf{CA} & \textbf{AA} & \textbf{CA} & \textbf{AA} & \textbf{CA} & \textbf{AA} & \textbf{CA} & \textbf{AA} \\
    \midrule
    \multirow{6}{*}{CIFAR-10}
    & Craig \citep{mirzasoleiman2020coresetsdataefficienttrainingmachine} 
        & 58.23 $\pm$ 1.68 & 44.49 $\pm$ 1.05 
        & 85.35 $\pm$ 0.44 & 52.90 $\pm$ 2.30 
        & 91.11 $\pm$ 0.33 & 46.28 $\pm$ 2.62 
        & 93.12 $\pm$ 0.13 & 48.51 $\pm$ 1.88 \\
    & kCenterGreedy \cite{sener2018activelearningconvolutionalneural} 
        & 50.46 $\pm$ 0.36 & 36.58 $\pm$ 0.25 
        & 85.26 $\pm$ 0.27 & 49.74 $\pm$ 1.67 
        & 91.98 $\pm$ 0.19 & 44.34 $\pm$ 2.51 
        & 93.92 $\pm$ 0.18 & 47.52 $\pm$ 1.32 \\
    & ModerateDS \cite{xia2023moderate} 
        & 59.34 $\pm$ 0.44 & 45.16 $\pm$ 0.62 
        & 84.86 $\pm$ 0.46 & 54.37 $\pm$ 3.06 
        & 91.18 $\pm$ 0.23 & 49.75 $\pm$ 1.32 
        & 93.24 $\pm$ 0.14 & 51.52 $\pm$ 2.29 \\
    & Uniform 
        & 58.73 $\pm$ 1.10 & 44.63 $\pm$ 0.84 
        & 84.56 $\pm$ 0.41 & \textbf{54.53 $\pm$ 2.09} 
        & 91.42 $\pm$ 0.25 & 49.82 $\pm$ 0.74 
        & 93.30 $\pm$ 0.24 & 49.51 $\pm$ 0.78 \\
    & \textbf{\sys (Ours)} 
        & 57.95 $\pm$ 0.83 & \textbf{46.66 $\pm$ 0.79} 
        & 77.40 $\pm$ 0.23 & 51.71 $\pm$ 1.14 
        & 85.02 $\pm$ 0.28 & \textbf{52.12 $\pm$ 0.47} 
        & 89.57 $\pm$ 0.17 & \textbf{56.11 $\pm$ 0.36} \\
    \midrule
    \multirow{6}{*}{CIFAR-100}
    & Craig \cite{mirzasoleiman2020coresetsdataefficienttrainingmachine} 
        & 21.07 $\pm$ 1.44 & 14.68 $\pm$ 1.09 
        & 53.13 $\pm$ 0.22 & 31.27 $\pm$ 1.20 
        & 68.06 $\pm$ 0.27 & 27.35 $\pm$ 0.63 
        & 73.61 $\pm$ 0.24 & 24.80 $\pm$ 0.64 \\
    & kCenterGreedy \cite{sener2018activelearningconvolutionalneural} 
        & 16.82 $\pm$ 0.53 & 11.46 $\pm$ 0.51 
        & 50.69 $\pm$ 2.28 & 27.86 $\pm$ 1.08 
        & 68.53 $\pm$ 0.24 & 24.37 $\pm$ 0.17 
        & 74.54 $\pm$ 0.36 & 23.77 $\pm$ 0.56 \\
    & ModerateDS \cite{xia2023moderate} 
        & 22.67 $\pm$ 0.26 & 15.69 $\pm$ 0.22 
        & 55.10 $\pm$ 0.47 & 25.77 $\pm$ 1.21 
        & 67.64 $\pm$ 0.12 & 24.42 $\pm$ 0.54 
        & 73.09 $\pm$ 0.26 & 24.22 $\pm$ 0.54 \\
    & Uniform 
        & 23.97 $\pm$ 0.82 & 16.53 $\pm$ 0.73 
        & 55.06 $\pm$ 0.64 & 29.33 $\pm$ 1.05 
        & 69.02 $\pm$ 0.19 & 25.78 $\pm$ 0.51 
        & 73.76 $\pm$ 0.14 & 24.45 $\pm$ 0.32 \\
    & \textbf{\sys (Ours)} 
        & 22.60 $\pm$ 0.69 & \textbf{18.86 $\pm$ 0.43} 
        & 53.12 $\pm$ 0.28 & \textbf{34.43 $\pm$ 1.21} 
        & 65.77 $\pm$ 0.22 & \textbf{33.27 $\pm$ 0.29} 
        & 72.01 $\pm$ 0.13 & \textbf{31.88 $\pm$ 0.14} \\
    \bottomrule
\end{tabular}
    }
\end{table*}

\subsection{Claim 3: Training on harder samples reduces robustness on easy samples}

\begin{figure}[h!]
\centering
\begin{subfigure}{0.3\textwidth}
    \includegraphics[width=\textwidth]{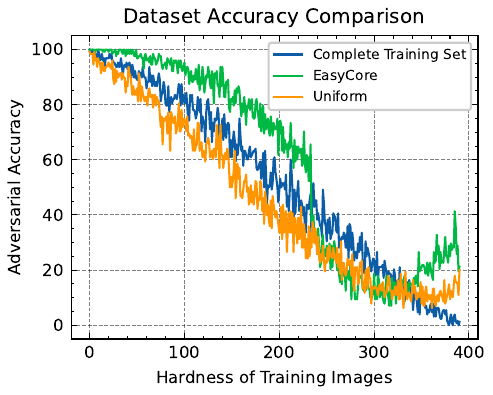}
    \caption{Training Images}
\end{subfigure}
\hfill
\begin{subfigure}{0.3\textwidth}
    \includegraphics[width=\textwidth]{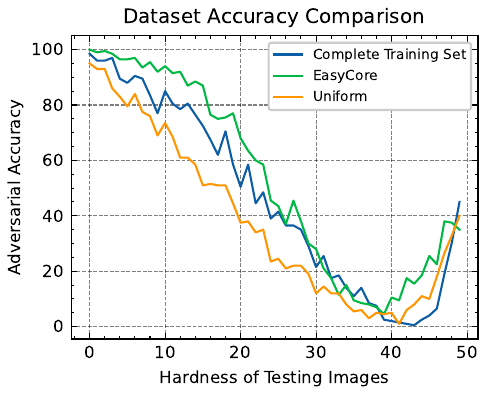}
    \caption{Testing Images}
\end{subfigure}
\caption{Adversarial Accuracy on images ordered by AIGN for both training and testing data for CIFAR-10.}
\label{fig:u_curve_ablation}
\end{figure}

We study the effect of sample hardness on adversarial accuracy for a fixed attack, within the training and testing dataset. We use the CIFAR-10 dataset for this study, using ResNet-18 models trained appropriately. We compare three training approaches - (i) standard training on the entire training dataset, (ii) standard training on a coreset fraction $0.6$ as chosen by \sys and (iii) standard training on a coreset fraction $0.6$ as chosen uniformly. First, we compute the list of indices ordered by ascending AIGN (refer \Cref{alg:EasyCore}). Then, we perform a batch-wise adversarial attack on these models for this fixed ordering of input images. The adversarial attack is done using a PGD-20 adversary with an attack parameter $\epsilon = \frac{8}{2550}$. The adversarial accuracies are plotted in \Cref{fig:u_curve_ablation}. We can see that \textit{harder} images have a significantly lower adversarial accuracy for a fixed attack in comparison with \textit{easier} images. On the training set, the adversarial accuracy linearly decreases with hardness. We also observe that \sys performs better in terms of adversarial accuracy on the lower AIGN images. This trend is observed both on the training and the test sets. This enhanced robustness obtained simply by omitting the \textit{harder} samples during training underpins the superior adversarial robustness of \sys.

\subsection{Claim 4: Adversarial training uses more dimensions to learn features}

We consider a similar visualization of clusters for an adversarially trained model as in a previous section. The model is trained using the TRADES approach~\cite{zhang2019trades}, using a training PGD adversary with attack parameter $\epsilon = \frac{8}{255}$. The result is shown in \Cref{fig:UMAP_plots} (c), (d). It can be seen that although the examples with lower AIGN appear in regions where it is easier to demarcate them, the distinction is much less clear than in \Cref{fig:UMAP_plots} (a), (b). This indicates that the feature space in the final layer of the model is not amenable to dimensionality reduction. To further substantiate this claim, we compute the minimum number of principal components $\kappa$ that, when combined, explain 95\% of the total variance in the $512$-dimension feature space in the penultimate layer. For a ResNet-18 model trained using standard training, $\kappa = 12$ whereas for an adversarially trained model trained against a PGD adversary with 
$\epsilon = \frac{8}{255}$, $\kappa = 143$. This stark difference in $\kappa$ can be interpreted as the tendency of adversarially trained models to \textit{consume} more of the parameter space of the model. This is in line with the claim made in \cite{madry2019deeplearningmodelsresistant} that adequate network capacity is required for adversarial training. 


\section{Results}
\label{sec:results}

In the subsequent sections, we compare the performance of \sys with existing coreset methods. We use CA to denote clean accuracy and AA to denote adversarial accuracy in Tables \ref{tab:accuracy_results_combined} - \ref{tab:imagenet_results}. 

\subsection{Comparative Benchmarking of \sys under Standard Training}

\begin{table*}[!htp]\centering
\caption{Accuracy under adversarial training for different methods and coreset fractions on CIFAR-10 and CIFAR-100. \sys shows up to a 5\% increase in clean and adversarial accuracy for the CIFAR-100 dataset with a coreset fraction of 20\% compared to uniform selection.}
\label{tab:adversarial_results}
\resizebox{\textwidth}{!}{
\begin{tabular}{lllcccccccc}
    \toprule
    \textbf{Dataset} & \textbf{Attack Parameter} & \textbf{Method}
    & \multicolumn{2}{c}{\textbf{0.05}} 
    & \multicolumn{2}{c}{\textbf{0.20}} 
    & \multicolumn{2}{c}{\textbf{0.40}} 
    & \multicolumn{2}{c}{\textbf{0.60}} \\
    \cmidrule(lr){4-5} \cmidrule(lr){6-7} \cmidrule(lr){8-9} \cmidrule(lr){10-11}
    & & & \textbf{CA} & \textbf{AA} & \textbf{CA} & \textbf{AA} & \textbf{CA} & \textbf{AA} & \textbf{CA} & \textbf{AA} \\
    \midrule
    \multirow{4}{*}{\textbf{CIFAR-10}} 
    & \multirow{2}{*}{$\epsilon = \frac{4}{255}$} 
    & \textbf{\sys (Ours)} 
        & 54.52 $\pm$ 0.75 & 21.83 $\pm$ 0.33 
        & 73.58 $\pm$ 0.50 & 27.80 $\pm$ 0.21 
        & 80.96 $\pm$ 0.49 & 32.94 $\pm$ 0.43 
        & 85.21 $\pm$ 0.11 & 37.31 $\pm$ 0.52 \\
    & 
    & Uniform 
        & 52.07 $\pm$ 2.00 & 16.01 $\pm$ 0.06 
        & 72.31 $\pm$ 0.21 & 26.73 $\pm$ 0.46 
        & 80.50 $\pm$ 0.28 & 32.91 $\pm$ 0.41 
        & 83.65 $\pm$ 0.21 & 36.21 $\pm$ 0.11 \\
    \cmidrule(lr){2-11}
    & \multirow{2}{*}{$\epsilon = \frac{8}{255}$} 
    & \textbf{\sys (Ours)} 
        & 48.77 $\pm$ 0.33 & 28.41 $\pm$ 0.36 
        & 68.05 $\pm$ 0.60 & 38.71 $\pm$ 0.13 
        & 75.68 $\pm$ 0.60 & 44.50 $\pm$ 0.29 
        & 79.45 $\pm$ 0.30 & 48.01 $\pm$ 0.24 \\
    & 
    & Uniform 
        & 50.12 $\pm$ 2.37 & 23.83 $\pm$ 0.26 
        & 66.94 $\pm$ 0.88 & 36.27 $\pm$ 0.46 
        & 74.77 $\pm$ 0.19 & 42.46 $\pm$ 0.23 
        & 78.07 $\pm$ 0.47 & 46.27 $\pm$ 0.24 \\
    \midrule
    \multirow{4}{*}{\textbf{CIFAR-100}} 
    & \multirow{2}{*}{$\epsilon = \frac{4}{255}$} 
    & \textbf{\sys (Ours)} 
        & 23.38 $\pm$ 0.39 & 10.02 $\pm$ 0.28 
        & 43.27 $\pm$ 0.40 & 16.95 $\pm$ 0.10 
        & 51.34 $\pm$ 0.22 & 19.94 $\pm$ 0.21 
        & 56.24 $\pm$ 0.21 & 21.37 $\pm$ 0.24 \\
    & 
    & Uniform 
        & 20.64 $\pm$ 0.43 & 5.01 $\pm$ 0.29 
        & 37.69 $\pm$ 0.42 & 10.76 $\pm$ 0.26 
        & 46.70 $\pm$ 0.33 & 14.59 $\pm$ 0.17 
        & 52.36 $\pm$ 0.53 & 17.28 $\pm$ 0.26 \\
    \cmidrule(lr){2-11}
    & \multirow{2}{*}{$\epsilon = \frac{8}{255}$} 
    & \textbf{\sys (Ours)} 
        & 22.23 $\pm$ 0.04 & 11.94 $\pm$ 0.13 
        & 39.12 $\pm$ 0.46 & 19.90 $\pm$ 0.15 
        & 46.32 $\pm$ 0.13 & 23.52 $\pm$ 0.02 
        & 50.25 $\pm$ 0.54 & 25.53 $\pm$ 0.14 \\
    & 
    & Uniform 
        & 19.84 $\pm$ 0.50 & 7.04 $\pm$ 0.40 
        & 34.72 $\pm$ 0.40 & 14.05 $\pm$ 0.38 
        & 42.95 $\pm$ 0.04 & 19.15 $\pm$ 0.33 
        & 47.54 $\pm$ 0.45 & 22.40 $\pm$ 0.08 \\
    \bottomrule
\end{tabular}

}
\end{table*}

\begin{table*}[!bhtp]\centering
\caption{Comparison with Adversarial Coreset Selection (ACS) for the ResNet-18 architecture with training attack parameter $\epsilon = \frac{8}{255}$. EasyCore performs better under higher coreset fractions and as the dataset complexity increases, while also consuming up to half the time at lower coreset fractions.}\label{tab:ACS_comparison}
\resizebox{\textwidth}{!}{ 
\begin{tabular}{llrrrrrrrrrrrrr}\toprule
\textbf{Dataset} &\textbf{Method/Fraction} &\multicolumn{3}{c}{\textbf{0.05}} &\multicolumn{3}{c}{\textbf{0.2}} &\multicolumn{3}{c}{\textbf{0.4}} &\multicolumn{3}{c}{\textbf{0.6}} \\\midrule
& &CA &AA &$\frac{T}{T_{EC}}$ &CA &AA &$\frac{T}{T_{EC}}$ &CA &AA &$\frac{T}{T_{EC}}$ &CA &AA &$\frac{T}{T_{EC}}$ \\
\midrule
\multirow{3}{*}{\centering \textbf{CIFAR-10}} &\textbf{\sys (Ours)} &48.77 &28.41 &1 &68.05 &38.71 &1 &75.68 &44.5 &1 &79.45 &48.01 &1 \\
&Adv. Craig (ACS-A) &53.06 &26.18 &1 &66.12 &36.05 &1 &71.52 &41.58 &1 &75.09 &44.27 &1 \\
&Adv. Craig (ACS-B) &56.54 &30.74 &$\approx 2$ &68.12 &40.26 &$\approx 1.26$ &73.12 &44.24 &$\approx 1.13$ &74.82 &45.98 &$\approx 1.08$ \\
\midrule
\multirow{3}{*}{\centering \textbf{CIFAR-100}} &\textbf{\sys (Ours)} &22.23 &11.94 &1 &39.12 &19.9 &1 &46.32 &23.52 &1 &50.25 &25.53 &1 \\
&Adv. Craig (ACS-A) &17.84 &5.54 &1 &30.73 &10.27 &1 &37.92 &13.45 &1 &42.09 &15.52 &1 \\
&Adv. Craig (ACS-B) &22.64 &7.81 &$\approx 2$ &36.48 &12.66 &$\approx 1.26$ &42.67 &15.7 &$\approx 1.13$ &46.45 &17.51 &$\approx 1.08$ \\
\bottomrule
\end{tabular}
}
\end{table*}

 For this comparison, we attack the model with a weaker PGD-20 adversary with attack parameter $\epsilon = \frac{8}{2550}$ for CIFAR-10 and CIFAR-100, and $\epsilon = \frac{4}{2550}$ on ImageNet-1K. We show superior adversarial accuracy in several coreset fractions in CIFAR-10 and \textbf{all} coreset fractions in CIFAR-100. The results are shown in \Cref{tab:accuracy_results_combined}. ImageNet-1K ~\cite{imagenet} is a relatively large dataset with 1000 classes. Using AIGN, we observe that not all classes are equally easy to learn. Therefore, we consider an augmented version of \sys wherein we balance the number of samples per class in addition to ordering based on AIGN. We show superior results in terms of adversarial accuracy compared to uniform sampling. The results are shown in \Cref{tab:imagenet_results}. 

\begin{table}[!htp]\centering
\caption{Accuracy details for standard training for ImageNet-1K and adversarial training for ImageNet-100. We show superior accuracy results across coreset fractions ranging, including as low as 5\%.}\label{tab:imagenet_results}
\resizebox{0.5\textwidth}{!}{ \begin{tabular}{p{0.2\textwidth}rrrrrrrr}
\toprule
\textbf{Dataset/Training} & \textbf{Method / Fraction} & \multicolumn{2}{c}{0.05} & \multicolumn{2}{c}{0.2} & \multicolumn{2}{c}{0.4} \\
\cmidrule(lr){3-4} \cmidrule(lr){5-6} \cmidrule(lr){7-8}
& & CA & AA & CA & AA & CA & AA \\
\midrule
\multirow{2}{0.2\textwidth}{ImageNet-1K (Standard Training)} 
  & Uniform & 35.23 & 15.39 & 61.56 & 17.83 & 69.42 & 16.11 \\
  & \textbf{EasyCore} & 37.86 & 18.01 & 58.41 & 21.14 & 66.93 & 19.46 \\
\cmidrule{1-8}
\multirow{2}{0.2\textwidth}{ImageNet-100 (Adversarial Training)} 
  & Uniform & 25.32 & 19.16 & 52.30 & 41.78 & 63.12 & 52.76 \\
  & \textbf{EasyCore} & 27.40 & 22.36 & 53.94 & 43.66 & 64.06 & 51.66 \\
\bottomrule
\end{tabular}
}
\end{table}

\subsection{Comparative Benchmarking of \sys under Adversarial Training}
We benchmark our method with adversarial training, with the TRADES approach ~\cite{zhang2019trades}. For the CIFAR-10 and CIFAR-100 datasets, we train using the ResNet18 model architecture. The model is attacked with a PGD-20 adversary with attack parameter $\epsilon = \frac{8}{255}$ (as per ~\cite{zhang2019trades}). We train with adversarial examples generated from a PGD-10 adversary, showing better adversarial accuracy on various training attack parameters (see \Cref{tab:adversarial_results}). For the ImageNet-100 dataset, we use the ResNet-50 architecture. We attack the model with a PGD-20 adversary with attack parameter $\epsilon = \frac{4}{255}$ (as per ~\cite{hadi}), having trained using the corresponding PGD-10 adversary. We continue to balance the number of samples per class, in order to offset the class imbalance. The results are shown in \Cref{tab:imagenet_results}. 

\subsection{Comparison with Adversarial Coreset Selection (ACS) \cite{hadi}}
\label{sec:ACS_comp}
The ACS algorithm as described in ~\cite{hadi} uses existing coreset approaches such as Craig \cite{mirzasoleiman2020coresetsdataefficienttrainingmachine} and GradMatch ~\cite{killamsetty2021gradmatchgradientmatchingbased}. The paper modifies the objective function to include an adversarial loss term to compute a coreset. The standard ACS approach consists of warm-starting the adversarial training using the entire dataset, thereafter switching to training on the coreset. However, \sys selects a coreset before training, without warm-start. Due to this difference in the coreset construction, we benchmark against the following two configurations of ACS, where warm-start is disabled. 

\begin{itemize}
    \item \textbf{ACS-A}: As a pre-processing step, we train a model for 30 epochs using full TRADES adversarial training, and use that to generate a coreset exactly once. Subsequently, we train a new model using that coreset using for 100 epochs using TRADES. 
    \item \textbf{ACS-B}: We obtain the initial coreset using the same pre-processing step as in ACS-A. Subsequently, we train for a total of 100 epochs, with a new coreset chosen every 20 epochs. This resembles the dynamic coreset approach specified in the paper. 
\end{itemize}

We also consider the time taken for the entire adversarial training with the coreset, ignoring the pre-processing steps, such as initial full adversarial training in ACS, or pre-training for \sys. We show this as the ratio between the time taken for ACS ($T$) and the time taken for \sys training ($T_{EC}$) for that coreset fraction. The results are shown in \Cref{tab:ACS_comparison}. 

It is important to note that the ACS-B employs a \textit{dynamic} selection strategy that resembles concepts of curriculum learning, as the coreset composition evolves through training. In contrast, \sys is a \textit{static} method where the coreset is selected once based on AIGN scores and remain fixed. We posit that this static nature is key to \sys's superior performance on more complex datasets (see CIFAR-100 results on \Cref{tab:ACS_comparison}). The dynamic re-selection in ACS-B allows \textit{hard} samples (high AIGN), which we identify as detrimental to robust generalization, to re-enter the training subset. As a dataset becomes more complex, the tails of its data distribution become longer, yielding harder samples \cite{feldman2021doeslearningrequirememorization} with higher AIGN scores. In contrast, \sys permanently removes these high-AIGN samples, enforces a consistent training curriculum on simpler and more prototypical examples, leading to a more robust final model.


\section{Conclusion}
\label{sec:conclusion}

In this paper, we propose \sys, an intuitively explainable yet quantitatively effective adversarial coreset algorithm using Average Input Gradient Norm (AIGN) over training. We demonstrate that the success of \sys is underpinned by the link between decision between AIGN, sample prototypical-ness and decision-boundary curvature. EasyCore uses a \textit{data-centric} approach to achieving robustness via adversarial coreset selection, thereby enhancing its efficiency and applicability. In order to demonstrate the efficacy of the data-centric approach, we benchmark the adversarial accuracy of \sys under standard training, where we outperform existing methods by up to 7\% in terms of adversarial accuracy. We also compare \sys under TRADES adversarial training, both against a uniform coreset and the ACS technique. We show improvement in adversarial accuracy of up to 5\% in this setting compared to existing methods. Subsequently, we present studies on the effect of training samples on the decision boundary between classes, and the nature of atypical and prototypical samples, thereby establishing the validity of our method across coreset fractions, datasets, models and training schemes.

\bibliography{aaai2026}

\appendix

\section{Appendix}
\label{Sec:Appendix}

\begin{figure}
\centering
    \includegraphics[width=0.3\textwidth]{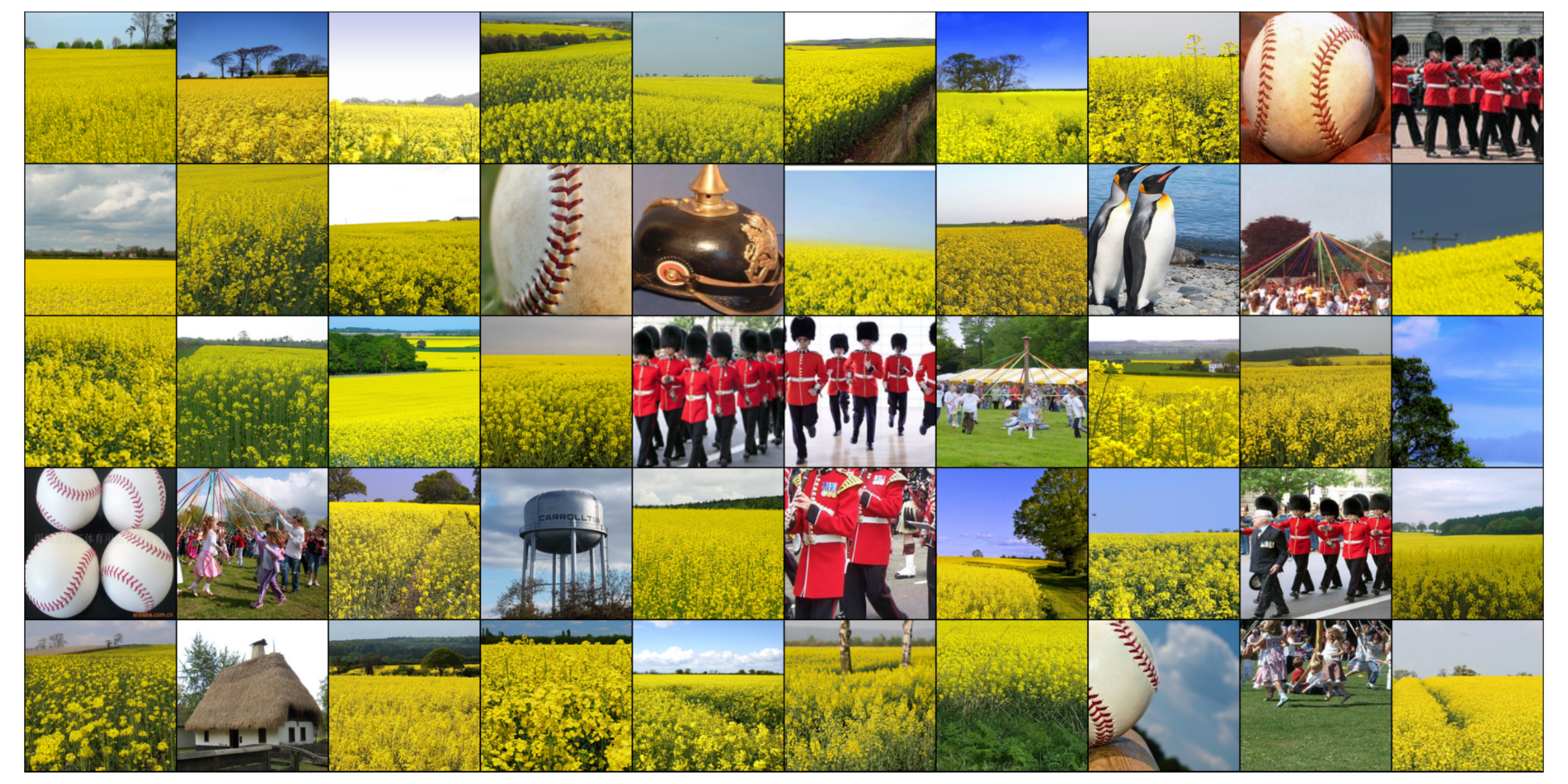}
    \caption{50 easiest images in ImageNet}
    \label{fig:50easy_imagenet}
\end{figure}

\subsection{Compute Used}

We used a single NVIDIA H200 GPU for both standard and adversarial training for CIFAR-10 and CIFAR-100, as well as adversarial training for ImageNet 100. We used a cluster of four NVIDIA H200 GPUs using Distributed Data Parallel to perform standard training on ImageNet. Our implementation uses the following libraries / frameworks: 

\begin{itemize}
    \item \textbf{Pytorch:} The overall framework for constructing datasets and models, as well as the implementation for the scheduler and optimizer used for training
    \item \textbf{Adversarial Robustness Toolbox:} The library used to create the required adversarial attacks and defenses used in our work
    \item \textbf{DeepCore:} The library used to construct coresets and benchmark with existing methods, under standard training.
\end{itemize}

Our implementation is attached along with this appendix in the supplementary section. 

\subsection{Hyperparameters used for various experiments}

We use the PyTorch library for all experiments. We outline the hyperparameters used in standard training in Table \Cref{tab:params_std} and those used for adversarial training in \Cref{tab:params_adv}

\begin{table*}[!htp]\centering
\caption{TRADES Adversarial Training Hyperparameters}\label{tab:params_adv}
\begin{tabular}{lrrrr}\toprule
\textbf{Hyperparameter} &\multicolumn{3}{c}{\textbf{Experiment}} \\\midrule
Dataset &CIFAR-10 &CIFAR-100 &ImageNet-100 \\
Model Arch. &ResNet-18 &ResNet-18 &ResNet-50 \\
Optimizer &SGD &SGD &SGD \\
Scheduler &MultiStepLR &MultiStepLR &MultiStepLR \\
Initial LR &0.1 &0.1 &0.1 \\
Training Epochs &100 &100 &100 \\
Momentum &0.9 &0.9 &0.9 \\
Weight Decay &2e-4 &2e-4 &2e-4 \\
Batch Size &128 &128 &128 \\
Visual Similarity Measure &$l_{\infty}$ &$l_{\infty}$ &$l_{\infty}$ \\
Attack Iterations &10 &10 &10 \\
Attack Parameter &$\epsilon \in \frac{\{4,8\}}{255}$ &$\epsilon \in \frac{\{4,8\}}{255}$ &$\epsilon = \frac{4}{255}$ \\
Attack Step Size &$\frac{1}{4} \cdot \epsilon$ &$\frac{1}{4} \cdot \epsilon$ &$\frac{1}{4} \cdot \epsilon$ \\
Number of experiments &3 &3 &1 \\
\bottomrule
\end{tabular}
\end{table*}

\begin{table*}[!htp]\centering
\caption{Standard Training Hyperparameters}\label{tab:params_std}
\begin{tabular}{lrrrr}\toprule
\textbf{Hyperparameter} &\multicolumn{3}{c}{\textbf{Experiment}} \\\midrule
Dataset &CIFAR-10 &CIFAR-100 &ImageNet \\
Model Arch. &ResNet-18 &ResNet-18 &ResNet-50 \\
Optimizer &SGD &SGD &SGD \\
Scheduler &CosineAnnealingLR &CosineAnnealingLR &StepLR \\
Initial LR &0.1 &0.1 &0.1 \\
Training Epochs &300 &300 &90 \\
Momentum &0.9 &0.9 &0.9 \\
Weight Decay &5e-4 &5e-4 &1e-4 \\
Batch Size &200 &200 &256 \\
Number of experiments &5 &5 &1 \\
\bottomrule
\end{tabular}
\end{table*}

For adversarial training using TRADES, we use the \textit{MultiStepLR()} scheduler with milestones at epochs 75 and 90, with a factor $\gamma = 0.1$. For standard training on the ImageNet dataset, we use the \textit{StepLR()} scheduler with a step size of 30 and $\gamma = 0.1$. 

\subsection{Details of dataset and model used in illustrative example for claim 2 (see \Cref{fig:DRI_vs_decision_boundary})}

The Python code describing the neural network used for training is shown below: \\

\begin{lstlisting}[language=Python]
class TenSkip2DNet(nn.Module):
    def __init__(self, hid=256, out=num_classes):
        super().__init__()
        self.input_layer = nn.Linear(2, hid)
        self.blocks = nn.ModuleList([
            nn.Sequential(nn.ReLU(), nn.Linear(hid, hid))
            for _ in range(20)
        ])
        self.head = nn.Linear(hid, out)
    def forward(self, x):
        x = self.input_layer(x)
        for blk in self.blocks:
            res = x
            x = blk(x)
            x = x + res
        return self.head(x)
\end{lstlisting}

The code used to generate the dataset is given below. \\

\begin{lstlisting}[language=Python]
num_classes, samples_per_class_train, samples_per_class_test, cluster_std = 2, [100,100,400,400,100,100], [250,250,1000,1000,250,250], [2.3, 2.3, 4.6, 4.6, 2.3, 2.3]
centers = [(-12, 0) , (-6, 0), (-6,-12), (0,12), (0,0), (6,0)]

X_train, y_train, X_test, y_test = [], [], [], []
for i, (cx, cy) in enumerate(centers):
    pts = np.random.randn(samples_per_class_train[i], 2) * cluster_std[i] + np.array([cx, cy])
    X_train.append(pts); y_train.append(np.full(samples_per_class_train[i], i%2))
    pts = np.random.randn(samples_per_class_test[i], 2) * cluster_std[i] + np.array([cx, cy])
    X_test.append(pts); y_test.append(np.full(samples_per_class_test[i], i%2))
X_train = np.vstack(X_train).astype(np.float32)
y_train = np.hstack(y_train).astype(np.int64)
X_test = np.vstack(X_test).astype(np.float32)
y_test = np.hstack(y_test).astype(np.int64)
\end{lstlisting}

\subsection{Related Mathematical Results}

We state and offer the proof of the link between input gradient and weight gradient, as stated in ~\cite{ravikumar2025towards}. 

\begin{lemma}[Input gradient norm is bound by weight gradient norm]
\label{lm:grad_w_x}
For any neural network, given a mini-batch of inputs $Z_b = (X_b, Y_b)$, the Frobenius norm of the gradient of the loss $\ell$ with respect to the input is bounded by the norm of the gradient with respect to the network's weights $\vec{w}_t$. Specifically:   
\begin{align}
\lVert\nabla_{X_b} \ell (\vec{w}_t, Z_b) \rVert_F &\leq  k_g \lVert\nabla_{w_t} \ell (\vec{w}_t, Z_b) \rVert_F
\end{align}where $k_g = \frac{\lVert W_t^{(1)} \rVert_F~ \lVert(X_b^\top)^+\rVert_F}{s_P}$ and  $s_P$ denotes the smallest singular value of $P =  X_b^\top (X_b^\top)^+$, where $^+$ denotes pseudo-inverse.
\end{lemma}
\textbf{Sketch of Proof.} The proof utilizes the chain rule to compute the gradients of the loss with respect to the inputs and the with respect to first-layer weights, establishing the relation between the two. By leveraging the Frobenius norm, the proof establishes an upper bound on the gradient with respect to the inputs in terms of the gradient with respect to the weights.

\subsection{Further Substantiation of Claim 4}

\begin{figure}[tb!]
\centering
    \includegraphics[width=0.4\textwidth]{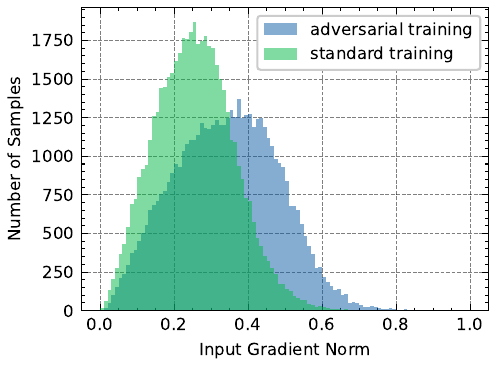}
    \caption{Comparison of average input gradient norm (normalized) for the CIFAR-100 dataset}
    \label{fig:density_comparsion}
\end{figure}

This claim is further substantiated by considering the distribution of the average input gradient norm for both the normally-trained and adversarially-trained models. We can observe that adversarial training shifts the mean of AIGN to the right, meaning that the number of samples with low AIGN reduces, and the cluster \textit{spreads out}. (\Cref{fig:density_comparsion})

\newpage

\end{document}